\documentclass{article}
\usepackage{times}
\usepackage{graphicx} 
\usepackage{subfigure} 
\usepackage{natbib}
\usepackage{algorithm}
\usepackage{algorithmic}
\usepackage{hyperref}

\usepackage[accepted]{icml2017} % After acceptance
\usepackage{amsmath}

\newcommand{\ignore}[1]{}

\icmltitlerunning{Cognitive Psychology for Deep Neural Networks: A Shape Bias Case Study}

\begin{document} 

\twocolumn[
\icmltitle{Cognitive Psychology for Deep Neural Networks: \\ A Shape Bias Case Study
}

\icmlsetsymbol{equal}{*}
\begin{icmlauthorlist}
\icmlauthor{Samuel Ritter}{equal,dm}
\icmlauthor{David G.T. Barrett}{equal,dm}
\icmlauthor{Adam Santoro}{dm}
\icmlauthor{Matt M. Botvinick}{dm}
\end{icmlauthorlist}
\icmlaffiliation{dm}{DeepMind, London, UK}
\icmlcorrespondingauthor{Samuel Ritter}{ritters@google.com}
\icmlcorrespondingauthor{David G.T. Barrett}{barrett@google.com}
\icmlkeywords{one-shot, word learning, bias, cognitive psychology, deep learning}
\vskip 0.3in
]

\printAffiliationsAndNotice{\icmlEqualContribution}

\begin{abstract} 
 Deep neural networks (DNNs) have achieved unprecedented performance on a wide range of complex tasks, rapidly outpacing our understanding of the nature of their solutions. This has caused a recent surge of interest in methods for rendering modern neural systems more interpretable. In this work, we propose to address the interpretability problem in modern DNNs using the rich history of problem descriptions, theories and experimental methods developed by cognitive psychologists to study the human mind. To explore the potential value of these tools, we chose a well-established analysis from developmental psychology that explains how children learn word labels for objects, and applied that analysis to DNNs. Using datasets of stimuli inspired by the original cognitive psychology experiments, we find that state-of-the-art one shot learning models trained on ImageNet exhibit a similar bias to that observed in humans: they prefer to categorize objects according to shape rather than color. The magnitude of this shape bias varies greatly among architecturally identical, but differently seeded models, and even fluctuates within seeds throughout training, despite nearly equivalent classification performance. These results demonstrate the capability of tools from cognitive psychology for exposing hidden computational properties of DNNs, while concurrently providing us with a computational model for human word learning.

\end{abstract}

\section{Introduction}

During the last half-decade deep learning has significantly improved performance on a variety of tasks (for a review, see~\citet{lecun2015deep}). However, deep neural network (DNN) solutions remain poorly understood, leaving many to think of these models as black boxes, and to question whether they can be understood at all~\citep{bornsteinartificial,lipton2016mythos}. This opacity obstructs both basic research seeking to improve these models, and applications of these models to real world problems~\citep{caruana2015intelligible}.

Recent pushes have aimed to better understand DNNs: tailor-made loss functions and architectures produce more interpretable features~\citep{higgins2016early,raposo2017discovering} while output-behavior analyses unveil previously opaque operations of these networks ~\citep{Karpathy2015visualizing}. Parallel to this work, neuroscience-inspired methods such as activation visualization~\citep{li2015visualizing},  ablation analysis~\citep{zeiler2014visualizing} and activation maximization~\citep{yosinski2015understanding} have also been applied. 

Altogether, this line of research developed a set of promising tools for understanding DNNs, each paper producing a glimmer of insight. Here, we propose another tool for the kit, leveraging methods inspired not by neuroscience, but instead by psychology. Cognitive psychologists have long wrestled with the problem of understanding another opaque intelligent system: the human mind. We contend that the search for a better understanding of DNNs may profit from the rich heritage of problem descriptions, theories, and experimental tools developed in cognitive psychology. To test this belief, we performed a proof-of-concept study on state-of-the-art DNNs that solve a particularly challenging task: one-shot word learning. Specifically, we investigate Matching Networks (MNs)~\citep{vinyals2016matching}, which have state-of-the-art one-shot learning performance on ImageNet and we investigate an Inception Baseline model~\citep{szegedy2015going}.

Following the approach used in cognitive psychology, we began by hypothesizing an inductive bias our model may use to solve a word learning task. Research in developmental psychology shows that when learning new words, humans tend to assign the same name to similarly shaped items rather than to items with similar color, texture, or size. To test the hypothesis that our DNNs discover this same ``shape bias'', we probed our models using datasets and an experimental setup based on the original shape bias studies \citep{landau1988importance}.

Our results are as follows: 1) Inception networks trained on ImageNet do indeed display a strong shape bias. 2) There is high variance in the bias between Inception networks initialized with different random seeds, demonstrating that otherwise identical networks converge to qualitatively different solutions. 3) MNs also have a strong shape bias, and this bias closely mimics the bias of the Inception model that provides input to the MN. 4) By emulating the shape bias observed in children, these models provide a candidate computational account for human one-shot word learning. Altogether, these results show that the technique of testing hypothesized biases using probe datasets can yield both expected and surprising insights about solutions discovered by trained DNNs.

\subsection{Related Work: Cognitive Modeling with Neural Networks}

The use of behavioral probes to understand neural network function has been extensively applied within psychology itself, where neural networks have been employed effectively as models of human cognitive function \citep{rumelhart1988parallel,plaut1996understanding,rogers2004semantic,mareschal2000connectionist}. In contrast, in the present work we are advocating for the application of behavioral probes along with associated theories and hypotheses from cognitive psychology to address the interpretability problem in modern deep networks. 

In spite of the widespread adoption of deep learning methods in recent years, to our knowledge, work applying behavioral probes to DNNs in machine learning for this purpose has been quite limited; we only are aware of \citet{zoran2015learning} and \citet{goodfellow2009measuring}, who used psychophysics-like experiments to better understand image processing models.
\section{Inductive Biases, Statistical Learners and Probe Datasets}

Before we delve into the specifics of the shape bias and one-shot word learning, we will describe our approach in the general context of inductive biases, probe datasets, and statistical learning. Suppose we have some data $\{y_i, x_i \}_{i=1}^N$ where $y_i=f(x_i)$. Our goal is to build a model of the data $g(.)$ to optimize some loss function $L$ measuring the disparity between $y$ and $g(x)$, e.g., $L=\sum_i ||y_i - g(x_i)||^2$. Perhaps this data $x$ is images of ImageNet objects to be classified, images and histology of tumors to be classified as benign or malignant \citep{kourou2015machine}, or medical history and vital measurements to be classified according to likely pneumonia outcomes \citep{caruana2015intelligible}. 

A statistical learner such as a DNN will minimize $L$ by discovering properties of the input $x$ that are predictive of the labels $y$. These discovered predictive properties are, in effect, the properties of $x$ for which the \textit{trained} model has an inductive bias. Examples of such properties include the shape of ImageNet objects, the number of nodes of a tumor, or a particular constellation of blood test values that often precedes an exacerbation of pneumonia symptoms.

Critically, in real-world datasets such as these, the discovered properties are unlikely to correspond to a single feature of the input $x$; instead they correspond to complex conjunctions of those features. We could describe one of these properties using a function $h(x)$, which, for example, returns the shape of the focal object given an ImageNet image, or the number of nodes given a scan of tumor. Indeed, one way to articulate the difficulty in understanding DNNs is to say that we often can't intuitively describe these conjunctions of features $h(x)$; although we often have numerical representations in intermediate DNN layers, they're often too arcane for us to interpret.

We advocate for addressing this problem using the following hypothesis-driven approach: First, propose a property $h_p(x)$ that the model may be using. Critically, it's not necessary that $h_p(x)$ be a function that can be evaluated using an automated method. Instead, the intention is that $h_p(x)$ is a function that humans (e.g. ML researchers and practitioners) can intuitively evaluate. $h_p(x)$ should be a property that is believed to be relevant to the problem, such as object shape or number of tumor nodes.

After proposing a property, the next step is to generate predictions about how the model should behave when given various inputs, if in fact it uses a bias with respect to the property $h_p(x)$. Then, construct and carry out an experiment wherein those predictions are tested. In order to execute such an experiment, it typically will be necessary to craft a set of probe examples $x$ that cover a relevant portion of the range of $h_p(x)$, for example a variety of object shapes. The results of this experiment will either support or fail to support the hypothesis that the model uses $h_p(x)$ to solve the task. This process can be especially valuable in situations where there is little or no training data available in important regions of the input space, and a practitioner needs to know how the trained model will behave in that region.  

Psychologists have developed a repertoire of such hypotheses and experiments in their effort to understand the human mind. Here we explore the application of one of these theory-experiment pairs to state of the art one-shot learning models. We will begin by describing the historical backdrop for the human one-shot word learning experiments that we will then apply to our DNNs.

\section{The problem of word learning; the solution of inductive biases}

Discussions of one-shot word learning in the psychological literature inevitably begin with the philosopher W.V.O. Quine, who broke this problem down and described one of its most computationally challenging components: there are an enormous number of tenable hypotheses that a learner can use to explain a single observed example. To make this point, Quine penned his now-famous parable of the field linguist who has gone to visit a culture whose language is entirely different from our own \citep{quine2013word}. The linguist is trying to learn some words from a helpful native, when a rabbit runs past. The native declares ``gavagai", and the linguist is left to infer the meaning of this new word. Quine points out that the linguist is faced with an abundance of possible inferences, including that ``gavagai" refers to rabbits, animals, white things, that specific rabbit, or ``undetached parts of rabbits". Quine argues that indeed there is an infinity of possible inferences to be made, and uses this conclusion to bolster the assertion that meaning itself cannot be defined in terms of internal mental events\footnotemark.

Contrary to Quine's intentions, when this example was introduced to the developmental psychology community by \citet{macnamara1972cognitive}, it spurred them not to give up on the idea of internal meaning, but instead to posit and test for cognitive biases that enable children to eliminate broad swaths of the hypothesis space \citep{bloom2000children}. A variety of hypothesis-eliminating biases were then proposed including the whole object bias, by which children assume that a word refers to an entire object and not its components \citep{markman1990constraints}; the taxonomic bias, by which children assume a word refers to the basic level category an object belongs to \citep{markman1984children}; the mutual exclusivity bias, by which children assume that a word only refers to one object category \citep{markman1988children}; the shape bias, with which we are concerned here \citep{landau1988importance}; and a variety of others \citep{bloom2000children}. These biases were tested empirically in experiments wherein children or adults were given an object (or picture of an object) along with a novel name, then were asked whether the name should apply to various other objects. 

Taken as a whole, this work yielded a computational level \citep{marr1982vision} account of word learning whereby people make use of biases to eliminate unlikely hypotheses when inferring the meaning of new words. Other contrasting and complementary approaches to explaining word learning exist in the psychological literature, including association learning \citep{regier1996human,colunga2005lexicon} and Bayesian inference \citep{xu2007word}. We leave the application of these theories to deep learning models to future work, and focus on determining what insight can be gained by applying a hypothesis elimination theory and methodology.

We begin the present work with the knowledge that part of the hypothesis elimination theory is correct: the models surely use some kind of inductive biases since they are statistical learning machines that successfully model the mapping between images and object labels. However, several questions remain open. What predictive properties did our DNNs find? Do all of them find the same properties? Are any of those properties interpretable to humans? Are they the same properties that children use? How do these biases change over the course of training?

To address these questions, we carry out experiments analogous to those of \citet{landau1988importance}. This enables us to test whether the shape bias -- a human interpretable feature used by children when learning language -- is visible in the behavior of MNs and Inception networks. Furthermore we are able to test whether these two models, as well as different instances of each of them, display the same bias. In the next section we will describe in detail the one-shot word learning problem, and the MNs and Inception networks we use to solve it.

\footnotetext{Unlike Quine, we use a pragmatic definition of meaning - a human or model understands the meaning of a word if they assign that word to new instances of objects in the correct category.}

\section{One-shot word learning models and training}
\subsection{One-shot word learning task}

The one-shot word learning task is to label a novel data example $\hat{x}$ (e.g. a novel probe image) with a novel class label $\hat{y}$  (e.g. a new word) after only a single example. More specifically, given a support set $S=\{(x_i,y_i): i \in [1,k]\}$, of images $x_i$ and their associated labels $y_i$, and an unlabelled probe image $\hat{x}$, the one-shot learning task is to identify the true label of the probe image $\hat{y}$ from the support set labels $\{y_i: i \in [1,k]\}$:
\begin{equation}
\label{eqn:one-shot-general}
    \hat{y} = \displaystyle{\arg \max_y P(y | \hat{x}, S)}  . 
\end{equation}
We assume that the image labels $y_i$ are represented using a one-hot encoding and that $ P(y | \hat{x}, S)$ is parameterised by a DNN, allowing us to leverage the ability of deep networks to learn powerful representations. 

\subsection{Inception: baseline one-shot learning model}

In our simplest \emph{baseline} one-shot architecture, a probe image $\hat{x}$ is given the label of the nearest neighbour from the support set: 
\begin{equation} 
\begin{split}
\hat{y} &= y \\ 
(x, y) &= \displaystyle{\arg \min_{(x_i,y_i) \in S } d(h(x_i),h(\hat{x}))}
\end{split}
\label{eqn:baseline}
\end{equation}
where $d$ is a distance function. The function $h$ is parameterised by Inception -- one of the best performing ImageNet classification models \citep{szegedy2015going}. Specifically, $h$ returns features from the last layer (the softmax input) of a pre-trained Inception classifier, where the Inception classifier is trained using rms-prop, as described in~\citet{szegedy2015rethinking}, section 8. With these features as input and cosine distance as the distance function, the classifier in equation~\ref{eqn:baseline} achieves 87.6\% accuracy on one-shot classification on the ImageNet dataset \citep{vinyals2016matching}. Henceforth, we call the Inception classifier together with the nearest-neighbor component the Inception Baseline (IB) model.

\subsection{Matching Nets model architecture and training}
\label{sec:MN}

We also investigate a state-of-the-art one-shot learning architecture called \emph{Matching Nets} (MN) \citep{vinyals2016matching}. MNs are a fully differentiable neural network architecture with state-of-the-art one shot learning performance on ImageNet (93.2\% one-shot labelling accuracy).  

MNs are trained to assign label $\hat{y}$ to probe image $\hat{x}$ according to equation \ref{eqn:one-shot-general} using an attention mechanism $a$ acting on image embeddings stored in  the support set $S$:
\begin{equation}
    a(\hat{x},x_i) = \frac{e^{d(f(\hat{x},S),g(x_i,S))}}{\sum_j e^{d(f(\hat{x},S),g(x_j,S)) }}, 
    \label{eqn:MNa}
\end{equation}
where $d$ is a cosine distance and where $f$ and $g$ provide context-dependent embeddings of $\hat{x}$ and $x_i$ (with context $S$). The embedding $g(x_i,S) $ is a bi-directional LSTM \citep{hochreiter1997long} with the support set $S$ provided as an input sequence. The embedding $f(\hat{x},S)$ is an LSTM with a read-attention mechanism operating over the entire embedded support set. The input to the LSTM is given by the penultimate layer features of a pre-trained deep convolutional network, specifically Inception, as in our baseline IB model described above \citep{szegedy2015going}. 

The training procedure for the one-shot learning task is critical if we want MNs to classify a probe image $\hat{x}$ after viewing only a single example of this new image class in its support set \citep{hochreiter2001learning,santoro2016meta}. 

To train MNs we proceed as follows: (1) At each step of training, the model is given a small support set of images and associated labels. In addition to the support set, the model is fed an unlabelled probe image $\hat{x}$; (2) The model parameters are then updated to improve classification accuracy of the probe image $\hat{x}$ given the support set. Parameters are updated using stochastic gradient descent with a learning rate of $0.1$; (3) After each update, the labels $\{y_i: i \in [1,k]\}$ in the training set are randomly re-assigned to new image classes (the label indices are randomly permuted, but the image labels are not changed). This is a critical step. It prevents MNs from learning a consistent mapping between a category and a label. Usually, in classification, this is what we want, but in one-shot learning we want to train our model for classification after viewing a single in-class example from the support set. Formally, our objective function is:
\begin{equation}
    L = E_{C\sim T}\left[ E_{S\sim C,B\sim C}\left[\displaystyle{\sum_{(x,y)\in B}\log P(y | x, S)}  \right]\right]
\end{equation}
where $T$ is the set of all possible labelings of our classes, $S$ is a support set sampled with a class labelling $C \sim T$ and $B$ is a batch of probe images and labels, also with the same randomly chosen class labelling as the support set. 

Next we will describe the probe datasets we used to test for the shape bias in the IB and MNs after ImageNet training.

\section{Data for bias discovery}
\label{sec:dataset}
\subsection{Cognitive Psychology Probe Data}

\begin{figure}[ht]
\begin{center}
\def \scalevar{0.12}
\centering
\begin{tabular}{ c c c c}
colour match &
shape match&
 &
probe

\\
  \includegraphics[scale=\scalevar]{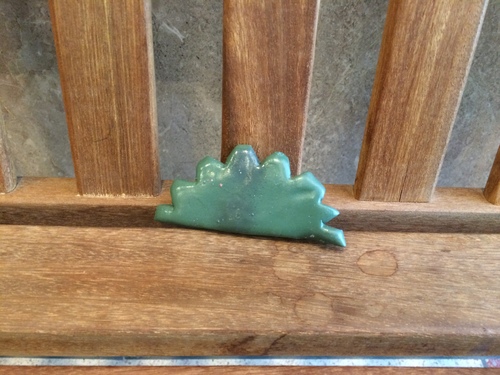}&
  \includegraphics[scale=\scalevar]{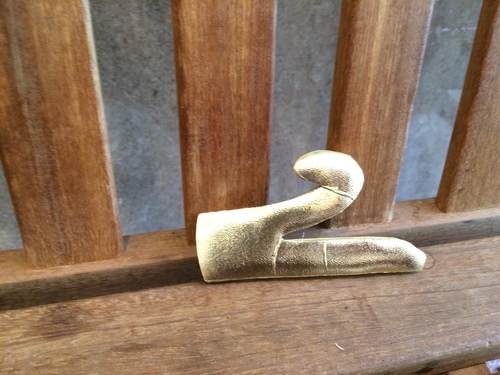}&  
   &
  \includegraphics[scale=\scalevar]{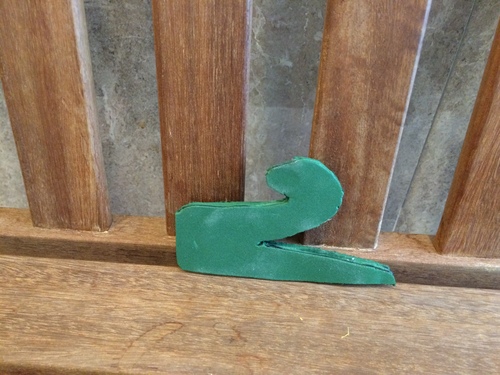}
    \\
  \includegraphics[scale=\scalevar]{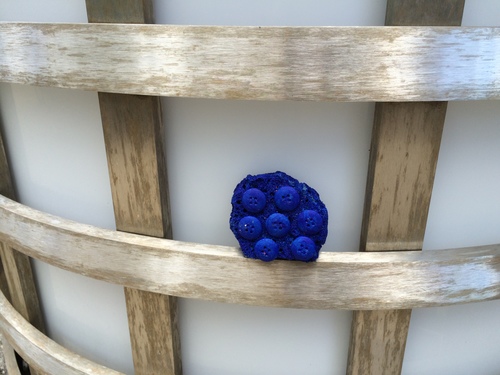} &
  \includegraphics[scale=\scalevar]{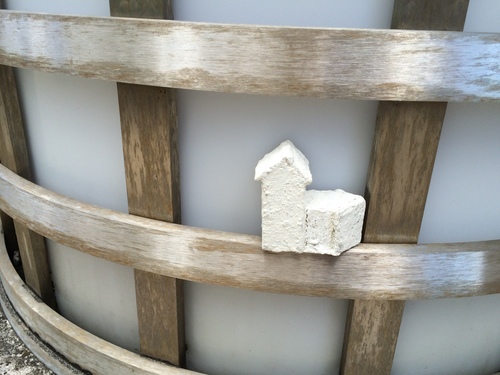}& 
   &
  \includegraphics[scale=\scalevar]{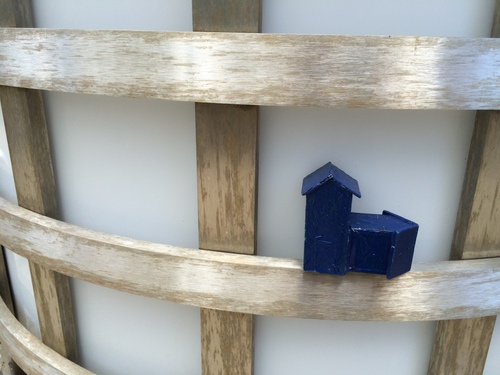}
  \\
  \includegraphics[scale=\scalevar]{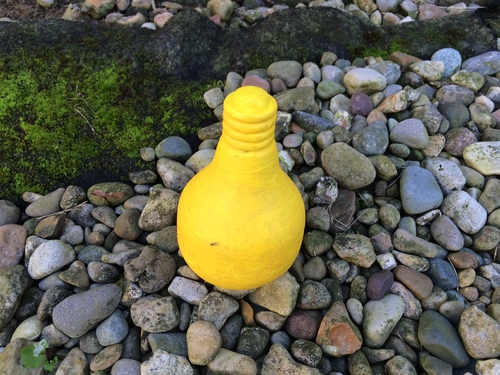}&
  \includegraphics[scale=\scalevar]{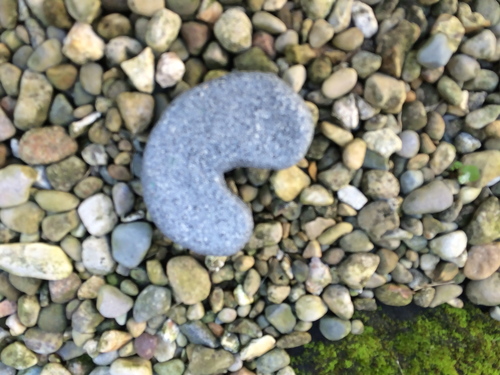}& 
   &
  \includegraphics[scale=\scalevar]{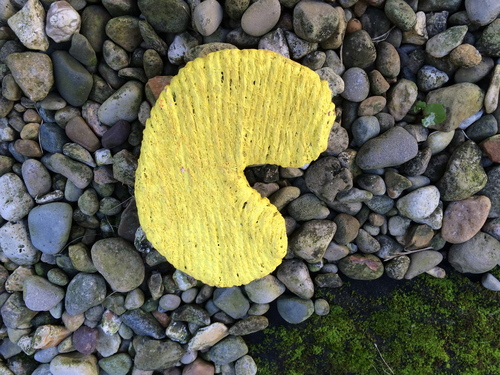}
    \\
  \includegraphics[scale=\scalevar]{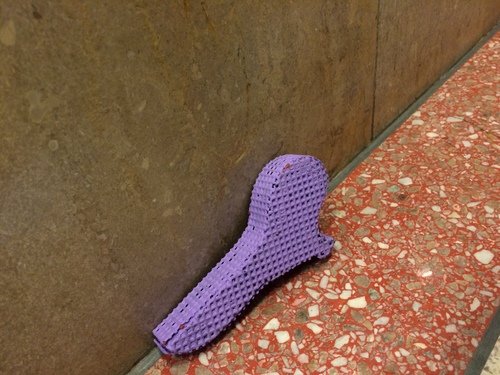} &
  \includegraphics[scale=\scalevar]{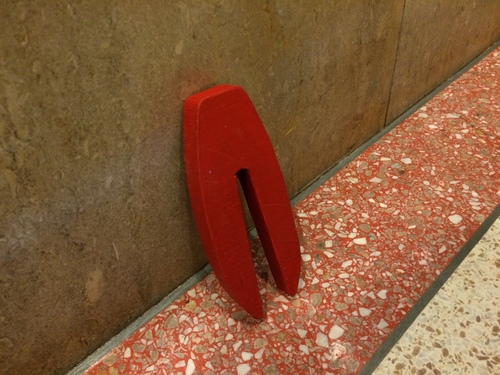}& 
   &
  \includegraphics[scale=\scalevar]{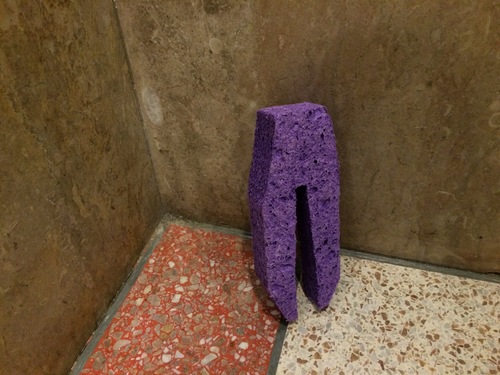}
\end{tabular}
\caption{Example images from the Cognitive Psychology Dataset (see section \ref{sec:dataset}). The data consists of image triples (rows), each containing a \emph{colour match} image (left column), a \emph{shape match} image (middle column) and a \emph{probe} image (right column). We use these triples to calculate the shape bias by reporting the proportion of times that a model assigns the shape match image class to the probe image. This dataset was supplied by cognitive psychologist Linda Smith, and was designed to control for object size and background.
  }
\label{fig:4_object_triples}
\end{center}
\vskip -0.2in
\end{figure}

The Cognitive Psychology Probe Data (CogPsyc data) that we use consists of 150 images of objects (Figure~\ref{fig:4_object_triples}). The images are arranged in triples consisting of a probe image, a shape-match image (that matches the probe in colour but not shape), and a color-match image (that matches the probe in shape but not colour). In the dataset there are 10 triples, each shown on 5 different backgrounds, giving a total of 50 triples.\footnotemark

\footnotetext{
The CogPsyc dataset is available at 
\url{http://www.indiana.edu/~cogdev/SB_testsets.html}
}

The images were generously provided by cognitive psychologist Linda Smith. The images are photographs of stimuli used previously in shape bias experiments conducted in the Cognitive Development Lab at Indiana University. The potentially confounding variables of background content and object size are controlled in this dataset.

\subsection{Probe Data from the wild}

We have also assembled a \emph{real-world} dataset consisting of 90 images of objects (30 triples) collected using Google Image Search. Again, the images are arranged in triples consisting of a probe, a shape-match and a colour-match. For the probe image, we chose images of real objects that are unlikely to appear in standard image datasets such as ImageNet. In this way, our data contains the irregularity of the real world while also probing our models' properties outside of the image space covered in our training data. For the shape-match image, we chose an object with a similar shape (but with a very different colour), and for the colour-match image, we chose an object with a similar colour (but with a very different shape). For example, one triple consists of a silver tuning fork as the probe, a silver guitar capo as the colour match, and a black tuning fork as the shape match. Each photo in the dataset  contains a single object on a white background.

We collected this data to strengthen our confidence in the results obtained for the CogPsych dataset and to demonstrate the ease with which such probe datasets can be constructed. One of the authors crafted this dataset solely using Google Image Search in the span of roughly two days' work. Our results with this dataset, especially the fact that the bias pattern over time matches the results from the well established CogPsych dataset, support the contention that DNN practitioners can collect effective probe datasets with minimal time expenditure using readily available tools.

\section{Results}

\begin{figure*}[ht]
\begin{center}
\centerline{\includegraphics[width=\textwidth]{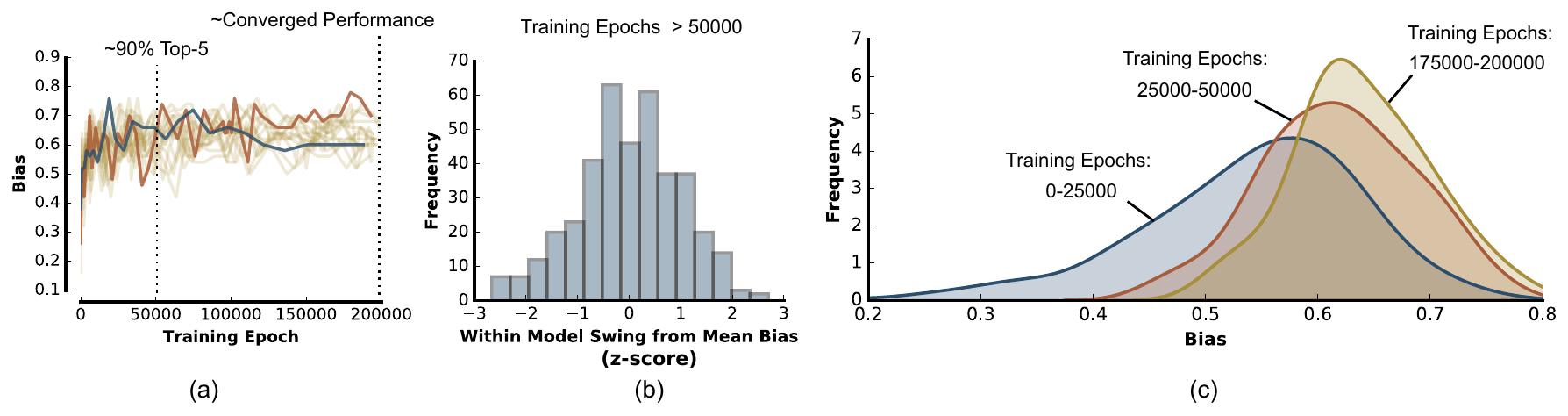}}
\caption{Shape bias across models with different initialization seeds, and within models during training calculated using the CogPsyc dataset. (a) The shape bias $B_s$ of 15 Inception models is calculated throughout training (yellow lines). A strong shape bias emerges across all models. A bias value $B_s > 0.5$ indicates a shape bias and $B_s < 0.5$ indicates a colour bias. Two examples are highlighted here (blue and red lines) for clarity. (b) The shape bias fluctuates strongly within models during training by up to three standard deviations. (c) The distribution of bias values, calculated at the start (blue), middle (red) and end (yellow) of training. Bias variability is high at the start and end of training. Here, these distributions are calculated using kernel density estimates from all shape bias measurements from all models within the indicated window.
}
\label{fig:statistics}
\end{center}
\vskip -0.2in
\end{figure*}

\begin{figure}[h]
\begin{center}
\centerline{\includegraphics[width=0.8\columnwidth]{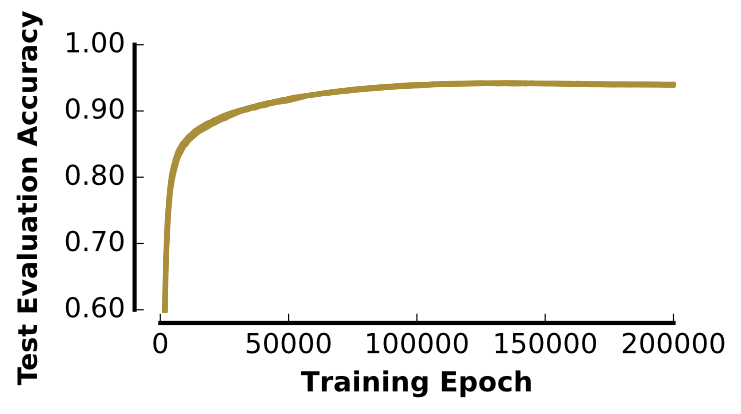}}
\caption{Classification accuracy of all 15 Inception models evaluated on a test set during training on ImageNet (same models as in Figure \ref{fig:statistics}) . All 15 Inception network seeds achieve near identical test accuracy (overlapping yellow lines).}
\label{fig:training}
\end{center}
\vskip -0.2in
\end{figure} 

\begin{figure}[h]
\begin{center}
\centerline{\includegraphics[width=0.7\columnwidth]{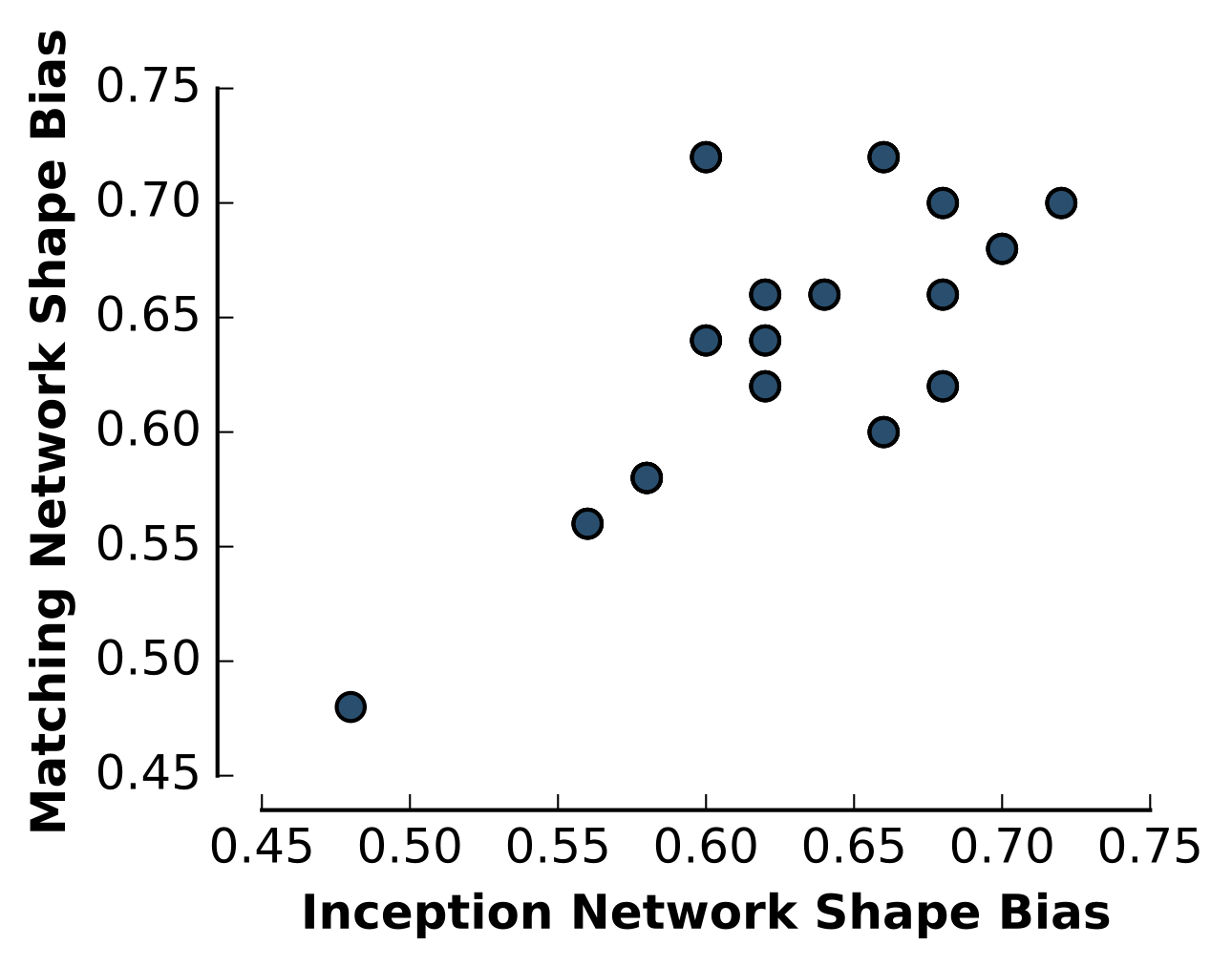}}
\caption{Scatter plot showing Matching Network (MN) bias as a function of Inception bias. Each MN receives input through an Inception model. Each point in this scatter plot is the bias of a MN and the bias of the Inception model providing input to that particular MN. In total, the bias values of 45 MN models are plotted (some dots are overlapping).}
\label{fig:mn_scatter}
\end{center}
\vskip -0.2in
\end{figure} 

\begin{figure*}[ht]
\begin{center}
\centerline{\includegraphics[width=\textwidth]{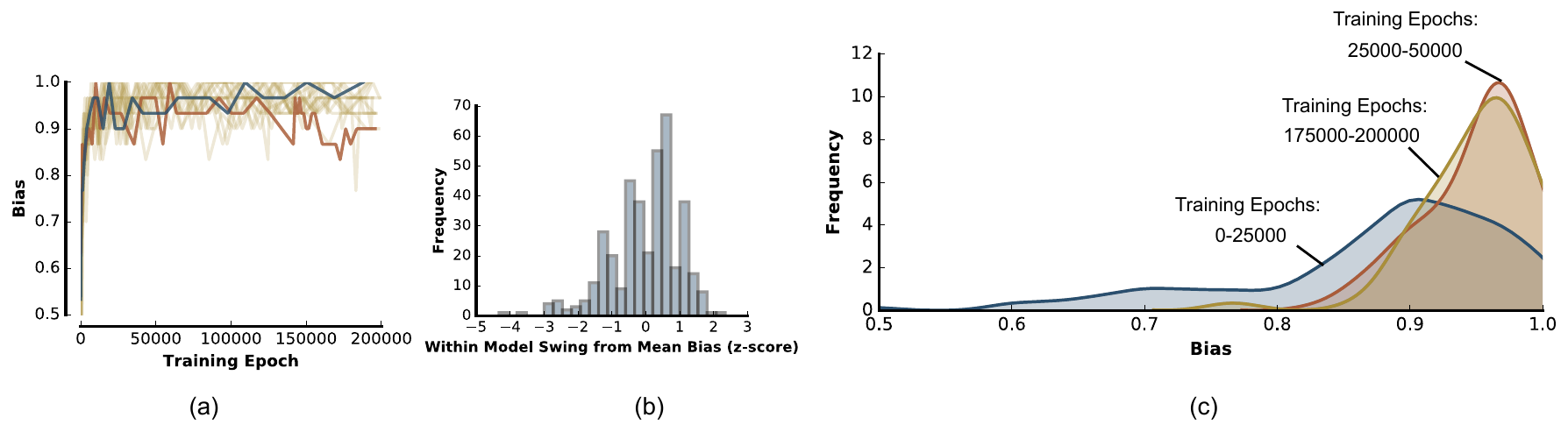}}
\caption{Shape bias across models with different initialization seeds, and within models during training calculated using the real-world dataset. (a) The shape bias $B_s$ of 15 Inception models is calculated throughout training (yellow lines). A strong shape bias emerges across all models. Two examples are highlighted here (blue and red lines) for clarity. (b) The shape bias fluctuates strongly within models during training. (c) The distribution of bias values, calculated at the start (blue), middle (red) and end (yellow) of training. Bias variability is high at the start and end of training. }
\label{fig:statistics_MN}
\end{center}
\vskip -0.2in
\end{figure*}

\subsection{Shape bias in the Inception Baseline Model}

First, we measured the shape bias in IB: we used a pre-trained Inception classifier (with 94\% top-5 accuracy) to provide features for our nearest-neighbour one-shot classifier, and probed the model using the CogPsyc dataset. Specifically, for a given probe image $\hat{x}$, we loaded the shape-match image $x_s$ and corresponding label $y_s$, along with the colour-match image $x_c$ and corresponding label $y_c$ into memory, as the support set $S = \{(x_s,y_s),(x_c,y_c)\}$. We then calculated $\hat{y}$ using Equation \ref{eqn:baseline}. Our model assigned either $y_c$ or $y_s$ to the probe image. To estimate the shape bias $B_s$, we calculated the proportion of shape labels assigned to the probe:
\begin{equation}
    B_s = \displaystyle{E(\delta(\hat{y} - y_s))},
\end{equation}
where $E$ is an expectation across probe images and $\delta$ is the Dirac delta function.

We ran all IB experiments using both Euclidean and cosine distance as the distance function. We found that the results for the two distance functions were qualitatively similar, so we only report results for Euclidean distance.

We found the shape bias of IB to be $B_s = 0.68$. Similarly, the shape bias of IB using our real-world dataset was $B_s = 0.97$. Together, these results strongly suggest that IB trained on ImageNet has a stronger bias towards shape than colour. 

Note that, as expected, the shape bias of this model is qualitatively similar across datasets while being quantitatively different - largely because the datasets themselves are quite different. Indeed, the datasets were chosen to be quite different so that we could explore a broad space of possibilities. In particular, our CogPsyc dataset backgrounds have much larger variability than our real-world dataset backgrounds, and our real-world dataset objects have much greater variability than the CogPsyc dataset objects. 

\subsection{Shape bias in the Matching Nets Model}
Next, we probed the MNs using a similar procedure. We used the IB trained in the previous section to provide the input features for the MN as described in section \ref{sec:MN}. Then, following the training procedure outlined in section \ref{sec:MN} we trained MNs for one-shot word learning on ImageNet, achieving state-of-the-art performance, as reported in \citep{vinyals2016matching}. Then, repeating the analysis above, we found that MNs have a shape of bias $B_s = 0.7$ using our CogPsyc dataset and a bias of $B_s = 1$ using the real-world dataset. It is interesting to note that these bias values are very similar to the IB bias values.

\subsection{Shape bias statistics: within models and across models}

The observation of a shape bias immediately raises some important questions. In particular: (1) Does this bias depend on the initial values of the parameters in our model? (2) Does the size of the shape bias depend on model performance? (3) When does shape bias emerge during training - before model convergence or afterwards? (4) How does shape bias compare between models, and within models?

To answer these questions, we extended the shape bias analysis described above to calculate the shape bias in a population of IB models and in a population of MN models with different random initialization (Figs. \ref{fig:statistics} and \ref{fig:statistics_MN}).

(1) We first calculated the dependence of shape bias on the initialization of IB (Fig. \ref{fig:statistics}). Surprisingly, we observed a strong variability, depending on the initialization. For the CogPsyc dataset, the average shape bias was $\overline{B}_s = 0.628$ with standard deviation $\sigma_{B_s} = 0.049$ at the end of training and for the real-world dataset the average shape bias was $\overline{B}_s = 0.958$ with $\sigma_{B_s} = 0.037$. 

(2) Next, we calculated the dependence of shape bias on model performance. For the CogPsych dataset, the correlation between bias and classification accuracy was $\rho=0.15$, with $t_{n=15}=0.55$, $p_{one\_tail}=0.29$, and for the \textit{real-world} dataset, the correlation was $\rho=-0.06$ with $t_{n=15}=-0.22$, $p_{one\_tail}=0.42$. Therefore, fluctuations in the bias cannot be accounted for by fluctuations in classification accuracy. This is not surprising, because the classification accuracy of all models was similar at the end of training, while the shape bias was variable. This demonstrates that models can have variable behaviour along important dimensions (e.g., bias) while having the same performance measured by another (e.g., accuracy). 

(3) Next we explored the emergence of the shape bias during training (Fig. \ref{fig:statistics}a,c; Fig. \ref{fig:statistics_MN}a,c). At the start of training, the average shape bias of these models was $\overline{B}_s = 0.448$ with standard deviation $\sigma_{B_s} = 0.0835$ on the CogPsyc dataset and $\overline{B}_s = 0.593$ with $\sigma_{B_s}= 0.073$ on the real-world dataset. We observe that a shape bias began to emerge very early during training, long before convergence. 

(4) Finally, we compare shape bias within models during training, and between models at the end of training. During training, the shape bias within IB fluctuates significantly (Fig. \ref{fig:statistics} b; Fig. \ref{fig:statistics_MN}b). In contrast, the shape bias does not fluctuate during training of the MN. Instead, the MN model inherits its shape bias characteristics at the start of training from the IB that provides it with input embeddings (Fig. \ref{fig:mn_scatter}) and this shape-bias remains constant throughout training. Moreover, there is no evidence that the MN and corresponding IB bias values are different from each other (paired t-test, $p = 0.167$). Note that we do not fine-tune the Inception model providing input while training the MN. We do this so that we can observe the shape-bias properties of the MN independent of the IB model properties.
\section{Discussion}

\subsection{A shape bias case study}

Our psychology-inspired approach to understanding DNNs produced a number of insights. Firstly, we found that both IB and MNs trained on ImageNet display a strong shape bias. This is an important result for practitioners who routinely use these models - especially for applications where it is known \emph{a priori} that colour is more important than shape. As an illustrative example, if a practitioner planned to build a one-shot fruit classification system, they should proceed with caution if they plan to use pre-trained ImageNet models like Inception and MNs because fruit are often defined according to colour features rather than shape. In applications where a shape bias is desirable (as is more often the case than not), this result provides reassurance that the models are behaving sensibly in the presence of ambiguity. 

The second surprising finding was the large variability in shape bias, both within models during training and across models, depending on the randomly chosen initialisation of our model. This variability can arise because our models are not being explicitly optimised for shape biased categorisation. This is an important result because it shows that not all models are created equally - some models will have a stronger preference for shape than others, even though they are architecturally identical and have almost identical classification accuracy. 

Our third finding -- that MNs retain the shape bias statistics of the downstream Inception network -- demonstrates the possibility for biases to propagate across model components. In this case, the shape bias propagates from the Inception model through to the MN memory modules. This result is yet another cautionary observation; when combining multiple modules together, we must be aware of contamination by unknown properties across modules. Indeed, a bias that is benign in one module might only have a detrimental effect when combined later with other modules.

A natural question immediately arises from these results - how can we remove an unwanted bias or induce a desirable bias? The biases under consideration are properties of an architecture and dataset synthesized together by an optimization procedure. As such, the observation of a shape-bias is partly a result of the statistics of natural image-labellings as captured in the ImageNet dataset, and partly a result of the architecture attempting to extract these statistics. Therefore, on discovering an unwanted bias, a practitioner can either attempt to change the model architecture to explicitly prevent the bias from emerging, or, they can attempt to manipulate the training data. If neither of these are possible - for example, if the appropriate data manipulation is too expensive, or, if the bias cannot be easily suppressed in the architecture, it may be possible to do zero-th order optimization of the models. For example, one may perform post-hoc model selection either using early stopping or by selecting a suitable model from the set of initial seeds.

An important caveat to note is that behavioral tools often do not provide insight into the neural mechanisms. In our case, the DNN mechanism whereby model parameters and input images interact to give rise to a shape bias have not been elucidated, nor did we expect this to happen. Indeed, just as cognitive psychology often does for neuroscience, our new computational level insights can provide a starting point for research at the mechanistic level. For example, in future work it would be interesting to use gradient-based visualization or neuron ablation techniques to augment the current results by identifying the mechanisms underlying the shape bias. The convergence of evidence from such introspective methods with the current behavioral method would create a richer account of these models' solutions to the one-shot word learning problem.

\subsection{Modelling human word learning}

There have been previous attempts to model human word learning in the cognitive science literature \citep{colunga2005lexicon,xu2007word,schilling2012taking,mayor2010neurocomputational}. However, none of these models are capable of one-shot word learning on the scale of real-world images. Because MNs both solve the task at scale and emulate hallmark experimental findings, we propose MNs as a computational-level account of human one-shot word learning. Another feature of our results supports this contention: in our model the shape bias increases dramatically early in training (Fig.~\ref{fig:statistics}a); similarly, humans show the shape bias much more strongly as adults than as children, and older children show the bias more strongly than younger children \citep{landau1988importance}.

As a good cognitive model should, our DNNs make testable predictions about word-learning in humans. Specifically, the current results predict that the shape bias should vary across subjects as well as within a subject over the course of development. They also predict that for humans with adult-level one-shot word learning abilities, there should be no correlation between shape bias magnitude and one-shot-word learning capability.

Another promising direction for future cognitive research would be to probe MNs for additional biases in order to predict novel computational properties in humans. Probing a model in this way is much faster than running human behavioural experiments, so a wider range of hypotheses for human word learning may be rapidly tested.

\subsection{Cognitive Psychology for Deep Neural Networks}
Through the one-shot learning case study, we demonstrated the utility of leveraging techniques from cognitive psychology for understanding the computational properties of DNNs. There is a wide ranging literature in cognitive psychology describing techniques for probing a spectrum of behaviours in humans. Our work here leads the way to the study of \emph{artificial cognitive psychology} - the application of these techniques to better understand DNNs. 

For example, it would be useful to apply work from the massive literature on episodic memory \citep{tulving1985elements} to the recent flurry of episodic memory architectures \citep{blundell2016model,graves2016hybrid}, and to apply techniques from the semantic cognition literature \citep{lamberts2013knowledge} to recent models of concept formation \citep{higgins2016early,gregor2016towards,raposo2017discovering}. More generally, the rich psychological literature will become increasingly useful for understanding deep reinforcement learning agents as they learn to solve increasingly complex tasks.

\section{Conclusion}

In this work, we have demonstrated how techniques from cognitive psychology can be leveraged to help us better understand DNNs. As a case study, we measured the shape bias in two powerful yet poorly understood DNNs - Inception and MNs. Our analysis revealed previously unknown properties of these models. More generally, our work leads the way for future exploration of DNNs using the rich body of techniques developed in cognitive psychology.

\section*{Acknowledgements} 

We would like to thank Linda Smith and Charlotte Wozniak for providing the Cognitive Psychology probe dataset; Charles Blundell for reviewing our paper prior to submission; Oriol Vinyals, Daan Wierstra, Peter Dayan, Daniel Zoran, Ian Osband and Karen Simonyan for helpful discussions; James Besley for legal assistance; and the DeepMind team for support.

\newpage
\bibliography{Cogpsyc}

\begin{thebibliography}{41}
\providecommand{\natexlab}[1]{#1}
\providecommand{\url}[1]{\texttt{#1}}
\expandafter\ifx\csname urlstyle\endcsname\relax
  \providecommand{\doi}[1]{doi: #1}\else
  \providecommand{\doi}{doi: \begingroup \urlstyle{rm}\Url}\fi

\bibitem[Bloom(2000)]{bloom2000children}
Bloom, Paul.
\newblock \emph{How children learn the meanings of words}.
\newblock MIT press Cambridge, MA, 2000.

\bibitem[Blundell et~al.(2016)Blundell, Uria, Pritzel, Li, Ruderman, Leibo,
  Rae, Wierstra, and Hassabis]{blundell2016model}
Blundell, Charles, Uria, Benigno, Pritzel, Alexander, Li, Yazhe, Ruderman,
  Avraham, Leibo, Joel~Z, Rae, Jack, Wierstra, Daan, and Hassabis, Demis.
\newblock Model-free episodic control.
\newblock \emph{arXiv preprint arXiv:1606.04460}, 2016.

\bibitem[Bornstein(2016)]{bornsteinartificial}
Bornstein, Aaron.
\newblock Is artificial intelligence permanently inscrutable? {D}espite new
  biology-like tools, some insist interpretation is impossible.
\newblock \emph{Nautilus}, 2016.

\bibitem[Caruana et~al.(2015)Caruana, Lou, Gehrke, Koch, Sturm, and
  Elhadad]{caruana2015intelligible}
Caruana, Rich, Lou, Yin, Gehrke, Johannes, Koch, Paul, Sturm, Marc, and
  Elhadad, Noemie.
\newblock Intelligible models for healthcare: Predicting pneumonia risk and
  hospital 30-day readmission.
\newblock In \emph{Proceedings of the 21th ACM SIGKDD International Conference
  on Knowledge Discovery and Data Mining}, pp.\  1721--1730. ACM, 2015.

\bibitem[Colunga \& Smith(2005)Colunga and Smith]{colunga2005lexicon}
Colunga, Eliana and Smith, Linda~B.
\newblock From the lexicon to expectations about kinds: a role for associative
  learning.
\newblock \emph{Psychological review}, 112\penalty0 (2):\penalty0 347, 2005.

\bibitem[Goodfellow et~al.(2009)Goodfellow, Lee, Le, Saxe, and
  Ng]{goodfellow2009measuring}
Goodfellow, Ian, Lee, Honglak, Le, Quoc~V, Saxe, Andrew, and Ng, Andrew~Y.
\newblock Measuring invariances in deep networks.
\newblock In \emph{Advances in neural information processing systems}, pp.\
  646--654, 2009.

\bibitem[Graves et~al.(2016)Graves, Wayne, Reynolds, Harley, Danihelka,
  Grabska-Barwi{\'n}ska, Colmenarejo, Grefenstette, Ramalho, Agapiou,
  et~al.]{graves2016hybrid}
Graves, Alex, Wayne, Greg, Reynolds, Malcolm, Harley, Tim, Danihelka, Ivo,
  Grabska-Barwi{\'n}ska, Agnieszka, Colmenarejo, Sergio~G{\'o}mez,
  Grefenstette, Edward, Ramalho, Tiago, Agapiou, John, et~al.
\newblock Hybrid computing using a neural network with dynamic external memory.
\newblock \emph{Nature}, 538\penalty0 (7626):\penalty0 471--476, 2016.

\bibitem[Gregor et~al.(2016)Gregor, Besse, Rezende, Danihelka, and
  Wierstra]{gregor2016towards}
Gregor, Karol, Besse, Frederic, Rezende, Danilo~Jimenez, Danihelka, Ivo, and
  Wierstra, Daan.
\newblock Towards conceptual compression.
\newblock In \emph{Advances In Neural Information Processing Systems}, pp.\
  3549--3557, 2016.

\bibitem[Higgins et~al.(2016)Higgins, Matthey, Glorot, Pal, Uria, Blundell,
  Mohamed, and Lerchner]{higgins2016early}
Higgins, Irina, Matthey, Loic, Glorot, Xavier, Pal, Arka, Uria, Benigno,
  Blundell, Charles, Mohamed, Shakir, and Lerchner, Alexander.
\newblock Early visual concept learning with unsupervised deep learning.
\newblock \emph{arXiv preprint arXiv:1606.05579}, 2016.

\bibitem[Hochreiter \& Schmidhuber(1997)Hochreiter and
  Schmidhuber]{hochreiter1997long}
Hochreiter, Sepp and Schmidhuber, J{\"u}rgen.
\newblock Long short-term memory.
\newblock \emph{Neural computation}, 9\penalty0 (8):\penalty0 1735--1780, 1997.

\bibitem[Hochreiter et~al.(2001)Hochreiter, Younger, and
  Conwell]{hochreiter2001learning}
Hochreiter, Sepp, Younger, A~Steven, and Conwell, Peter~R.
\newblock Learning to learn using gradient descent.
\newblock In \emph{International Conference on Artificial Neural Networks},
  pp.\  87--94, 2001.

\bibitem[Karpathy et~al.(2015)Karpathy, Johnson, and
  Fei-Fei]{Karpathy2015visualizing}
Karpathy, Andrej, Johnson, Justin, and Fei-Fei, Li.
\newblock Visualizing and understanding recurrent networks.
\newblock \emph{arXiv preprint arXiv:1506.02078}, 2015.

\bibitem[Kourou et~al.(2015)Kourou, Exarchos, Exarchos, Karamouzis, and
  Fotiadis]{kourou2015machine}
Kourou, Konstantina, Exarchos, Themis~P, Exarchos, Konstantinos~P, Karamouzis,
  Michalis~V, and Fotiadis, Dimitrios~I.
\newblock Machine learning applications in cancer prognosis and prediction.
\newblock \emph{Computational and structural biotechnology journal},
  13:\penalty0 8--17, 2015.

\bibitem[Lamberts \& Shanks(2013)Lamberts and Shanks]{lamberts2013knowledge}
Lamberts, Koen and Shanks, David.
\newblock \emph{Knowledge Concepts and Categories}.
\newblock Psychology Press, 2013.

\bibitem[Landau et~al.(1988)Landau, Smith, and Jones]{landau1988importance}
Landau, Barbara, Smith, Linda~B, and Jones, Susan~S.
\newblock The importance of shape in early lexical learning.
\newblock \emph{Cognitive development}, 3\penalty0 (3):\penalty0 299--321,
  1988.

\bibitem[LeCun et~al.(2015)LeCun, Bengio, and Hinton]{lecun2015deep}
LeCun, Yann, Bengio, Yoshua, and Hinton, Geoffrey.
\newblock Deep learning.
\newblock \emph{Nature}, 521\penalty0 (7553):\penalty0 436--444, 2015.

\bibitem[Li et~al.(2015)Li, Chen, Hovy, and Jurafsky]{li2015visualizing}
Li, Jiwei, Chen, Xinlei, Hovy, Eduard, and Jurafsky, Dan.
\newblock Visualizing and understanding neural models in nlp.
\newblock \emph{arXiv preprint arXiv:1506.01066}, 2015.

\bibitem[Lipton(2016)]{lipton2016mythos}
Lipton, Zachary~C.
\newblock The mythos of model interpretability.
\newblock \emph{arXiv preprint arXiv:1606.03490}, 2016.

\bibitem[Macnamara(1972)]{macnamara1972cognitive}
Macnamara, John.
\newblock Cognitive basis of language learning in infants.
\newblock \emph{Psychological review}, 79\penalty0 (1):\penalty0 1, 1972.

\bibitem[Mareschal et~al.(2000)Mareschal, French, and
  Quinn]{mareschal2000connectionist}
Mareschal, Denis, French, Robert~M, and Quinn, Paul~C.
\newblock A connectionist account of asymmetric category learning in early
  infancy.
\newblock \emph{Developmental psychology}, 2000.

\bibitem[Markman(1990)]{markman1990constraints}
Markman, Ellen~M.
\newblock Constraints children place on word meanings.
\newblock \emph{Cognitive Science}, 14\penalty0 (1):\penalty0 57--77, 1990.

\bibitem[Markman \& Hutchinson(1984)Markman and
  Hutchinson]{markman1984children}
Markman, Ellen~M and Hutchinson, Jean~E.
\newblock Children's sensitivity to constraints on word meaning: Taxonomic
  versus thematic relations.
\newblock \emph{Cognitive psychology}, 16\penalty0 (1):\penalty0 1--27, 1984.

\bibitem[Markman \& Wachtel(1988)Markman and Wachtel]{markman1988children}
Markman, Ellen~M and Wachtel, Gwyn~F.
\newblock Children's use of mutual exclusivity to constrain the meanings of
  words.
\newblock \emph{Cognitive psychology}, 20\penalty0 (2):\penalty0 121--157,
  1988.

\bibitem[Marr(1982)]{marr1982vision}
Marr, David.
\newblock Vision: A computational investigation into the human representation
  and processing of visual information, henry holt and co.
\newblock \emph{Inc., New York, NY}, 2:\penalty0 4--2, 1982.

\bibitem[Mayor \& Plunkett(2010)Mayor and
  Plunkett]{mayor2010neurocomputational}
Mayor, Julien and Plunkett, Kim.
\newblock A neurocomputational account of taxonomic responding and fast mapping
  in early word learning.
\newblock \emph{Psychological review}, 117\penalty0 (1):\penalty0 1, 2010.

\bibitem[Plaut et~al.(1996)Plaut, McClelland, Seidenberg, and
  Patterson]{plaut1996understanding}
Plaut, David~C, McClelland, James~L, Seidenberg, Mark~S, and Patterson,
  Karalyn.
\newblock Understanding normal and impaired word reading: computational
  principles in quasi-regular domains.
\newblock \emph{Psychological review}, 103\penalty0 (1):\penalty0 56, 1996.

\bibitem[Quine(1960)]{quine2013word}
Quine, Willard Van~Orman.
\newblock \emph{Word and object}.
\newblock MIT press, 1960.

\bibitem[Raposo et~al.(2017)Raposo, Santoro, Barrett, Pascanu, Lillicrap, and
  Battaglia]{raposo2017discovering}
Raposo, David, Santoro, Adam, Barrett, David~G.T., Pascanu, Razvan, Lillicrap,
  Timothy, and Battaglia, Peter.
\newblock Discovering objects and their relations from entangled scene
  representations.
\newblock \emph{arXiv preprint arXiv:1702.05068}, 2017.

\bibitem[Regier(1996)]{regier1996human}
Regier, Terry.
\newblock \emph{The human semantic potential: Spatial language and constrained
  connectionism}.
\newblock MIT Press, 1996.

\bibitem[Rogers \& McClelland(2004)Rogers and McClelland]{rogers2004semantic}
Rogers, Timothy~T and McClelland, James~L.
\newblock \emph{Semantic cognition: A parallel distributed processing
  approach}.
\newblock MIT press, 2004.

\bibitem[Rumelhart et~al.(1988)Rumelhart, McClelland, Group,
  et~al.]{rumelhart1988parallel}
Rumelhart, David~E, McClelland, James~L, Group, PDP~Research, et~al.
\newblock \emph{Parallel distributed processing}, volume~1.
\newblock IEEE, 1988.

\bibitem[Santoro et~al.(2016)Santoro, Bartunov, Botvinick, Wierstra, and
  Lillicrap]{santoro2016meta}
Santoro, Adam, Bartunov, Sergey, Botvinick, Matthew, Wierstra, Daan, and
  Lillicrap, Timothy.
\newblock Meta-learning with memory-augmented neural networks.
\newblock In \emph{Proceedings of The 33rd International Conference on Machine
  Learning}, pp.\  1842--1850, 2016.

\bibitem[Schilling et~al.(2012)Schilling, Sims, and
  Colunga]{schilling2012taking}
Schilling, Savannah~M, Sims, Clare~E, and Colunga, Eliana.
\newblock Taking development seriously: Modeling the interactions in the
  emergence of different word learning biases.
\newblock In \emph{CogSci}, 2012.

\bibitem[Szegedy et~al.(2015{\natexlab{a}})Szegedy, Liu, Jia, Sermanet, Reed,
  Anguelov, Erhan, Vanhoucke, and Rabinovich]{szegedy2015going}
Szegedy, Christian, Liu, Wei, Jia, Yangqing, Sermanet, Pierre, Reed, Scott,
  Anguelov, Dragomir, Erhan, Dumitru, Vanhoucke, Vincent, and Rabinovich,
  Andrew.
\newblock Going deeper with convolutions.
\newblock In \emph{Proceedings of the IEEE Conference on Computer Vision and
  Pattern Recognition}, pp.\  1--9, 2015{\natexlab{a}}.

\bibitem[Szegedy et~al.(2015{\natexlab{b}})Szegedy, Vanhoucke, Ioffe, Shlens,
  and Wojna]{szegedy2015rethinking}
Szegedy, Christian, Vanhoucke, Vincent, Ioffe, Sergey, Shlens, Jonathon, and
  Wojna, Zbigniew.
\newblock Rethinking the inception architecture for computer vision.
\newblock \emph{arXiv preprint arXiv:1512.00567}, 2015{\natexlab{b}}.

\bibitem[Tulving(1985)]{tulving1985elements}
Tulving, Endel.
\newblock Elements of episodic memory.
\newblock 1985.

\bibitem[Vinyals et~al.(2016)Vinyals, Blundell, Lillicrap, Kavukcuoglu, and
  Wierstra]{vinyals2016matching}
Vinyals, Oriol, Blundell, Charles, Lillicrap, Timothy, Kavukcuoglu, Koray, and
  Wierstra, Daan.
\newblock Matching networks for one shot learning.
\newblock \emph{arXiv preprint arXiv:1606.04080}, 2016.

\bibitem[Xu \& Tenenbaum(2007)Xu and Tenenbaum]{xu2007word}
Xu, Fei and Tenenbaum, Joshua~B.
\newblock Word learning as bayesian inference.
\newblock \emph{Psychological review}, 114\penalty0 (2):\penalty0 245, 2007.

\bibitem[Yosinski et~al.(2015)Yosinski, Clune, Nguyen, Fuchs, and
  Lipson]{yosinski2015understanding}
Yosinski, Jason, Clune, Jeff, Nguyen, Anh, Fuchs, Thomas, and Lipson, Hod.
\newblock Understanding neural networks through deep visualization.
\newblock \emph{arXiv preprint arXiv:1506.06579}, 2015.

\bibitem[Zeiler \& Fergus(2014)Zeiler and Fergus]{zeiler2014visualizing}
Zeiler, Matthew~D and Fergus, Rob.
\newblock Visualizing and understanding convolutional networks.
\newblock In \emph{European Conference on Computer Vision}, pp.\  818--833,
  2014.

\bibitem[Zoran et~al.(2015)Zoran, Isola, Krishnan, and
  Freeman]{zoran2015learning}
Zoran, Daniel, Isola, Phillip, Krishnan, Dilip, and Freeman, William~T.
\newblock Learning ordinal relationships for mid-level vision.
\newblock In \emph{Proceedings of the IEEE International Conference on Computer
  Vision}, pp.\  388--396, 2015.

\end{thebibliography}
\bibliographystyle{icml2017}

\end{document}